\documentclass[runningheads]{llncs}
\usepackage[T1]{fontenc}
\usepackage{amsmath}
\usepackage{amssymb}
\usepackage{graphicx}
\usepackage{pifont}
\usepackage{booktabs} 
\usepackage{multirow}
\usepackage{array}    
\usepackage{hyperref}

\usepackage{color}

\begin{document}
\title{Select, Don’t Train: The Benefits of Modular Entity Disambiguation with LLM-Based Selection}
\titlerunning{The Benefits of Modular Entity Disambiguation}

\author{Fina Polat\inst{1} \and Daniel Daza\inst{2} \and Pengyu Zhang\inst{1} \and Klim Zaporojets\inst{3} \and Paul Groth\inst{1}}

\authorrunning{Polat et al.}

\institute{
University of Amsterdam\\
\email{\{f.yilmazpolat,p.zhang,p.t.groth\}@uva.nl}
\and
Vrije Universiteit Amsterdam\\
\email{d.dazacruz@vu.nl}
\and
Aarhus University\\
\email{klim@cs.au.dk}
}

\maketitle

\begin{abstract}
Entity Disambiguation (ED) is a key task for constructing and using knowledge graphs. State-of-the-art neural approaches commonly model ED as a single task, although it consists of two distinct subproblems: retrieving candidate entities and selecting the correct one given context. Dual-encoder models optimize for both within a shared embedding space, forcing representations to balance high-recall retrieval with fine-grained selection, and they require trained retrievers, which are costly to maintain as knowledge graphs change. While recent work has begun to combine retrievers with LLM-based selectors, the interplay between the two stages has not been studied systematically. In this paper, we present a systematic comparison of retrieval strategies for candidate generation under a shared LLM-based selection stage, combining sparse retrieval (BM25), Web KB search, and a state-of-the-art trained dense retriever with several open- and closed-source LLMs. We show that, once selection is delegated to a capable LLM, training the retriever provides only modest additional value: a fully training-free BM25 retriever paired with an LLM selector reaches a new state of the art on the ZELDA benchmark, raising inKB micro-F1 from 82.3 to 86.3 (+4); pairing the same LLM with a trained dense retriever reaches 88.5. Decoupling retrieval from selection also exposes a limitation of current ED systems: when the correct entity is missing from retrieved candidates, they are forced to predict an incorrect entity. In contrast, our framework allows for abstention when retrieval failure is detected. In an evaluation setting that rewards correct abstentions, the training-free BM25 + LLM pipeline reaches 90.7 F1.
\end{abstract}

\section{Introduction}

Entity Disambiguation (ED) maps ambiguous textual mentions to entries in a knowledge base (KB), and is an important component of knowledge graph construction, information extraction, and question answering \cite{hoffart-etal-2011-robust,martinez2020information}. A common formulation of ED decomposes the task into two subproblems: retrieving a set of candidate entities for a mention, and selecting the correct entity given the surrounding context \cite{bunescu-pasca-2006-using}. Recent neural approaches, however, often model these steps jointly through shared embedding spaces that support both candidate retrieval and entity selection \cite{gillick-etal-2019-learning,wang-etal-2024-entity,ayoola-etal-2022-refined}.

This joint formulation has several practical and methodological implications. Retrieval and selection place different demands on representations: retrieval requires broad coverage over large entity collections, whereas selection depends on fine-grained contextual discrimination among semantically similar candidates. In addition, dense retrieval components are typically tied to trained encoders and precomputed indexes, making them comparatively costly to maintain as KBs evolve.

At the same time, recent progress in Large Language Models (LLMs) has led to renewed interest in modular ED pipelines in which retrieval and selection are treated as separate components \cite{ding-etal-2024-chatel,xiao-etal-2023-instructed,zhang-etal-2024-onegen}. LLMs provide strong in-context learning, reasoning, and instruction following capabilities and can be used to select among retrieved candidates without requiring task-specific training \cite{few_shot_paper,cot_paper,sun-etal-2023-chatgpt,instruct_gpt_paper}. This raises a broader question about the role of retrieval in modern ED systems: once contextual selection is delegated to an LLM, what properties are required from the retriever?

In this paper, we study this question through a systematic comparison of retrieval-based candidate generation strategies when using an LLM-based selection stage. We consider sparse retrieval, dense retrieval, and Web KB search interfaces, and analyze how these retrieval methods interact with LLM-based selection across several ED datasets. This modular setting also makes it possible to analyze retrieval failure independently from selection behavior. We additionally examine abstention-aware evaluation settings \cite{van-erp-etal-2016-evaluating,zhu-etal-2023-learn,zhou-etal-2024-gendecider}, where the model may reject the candidate set when retrieval appears to have failed, rather than being forced to predict an entity.

We evaluate on the ZELDA benchmark suite \cite{milich-akbik-2023-zelda}, which includes AIDA \cite{aida_paper}, WNED-WIKI and WNED-CWEB \cite{wned_cweb_paper}, ShadowLinks \cite{provatorova-etal-2021-robustness}, and the social-media datasets REDDIT \cite{botzer2021reddit} and TWEEKI \cite{harandizadeh-singh-2020-tweeki}. Our results suggest that lightweight retrieval approaches can remain competitive when combined with LLM-based selection. In particular, training-free retrieval methods such as BM25 achieve strong performance while avoiding retraining costs associated with dense retrieval models. We further find that abstention-aware evaluation provides additional insight into retrieval failure and selection behavior in modular ED systems.

Our main contributions are:
\begin{enumerate}
\item A systematic analysis of retrieval and selection in LLM-based ED pipelines, comparing sparse, dense, and Web-based retrieval strategies under a shared LLM selection stage.
\item Experiments with training-free LLM-based selection using both closed and open-source LLMs. We additionally explore low-resource fine-tuning of smaller 8B and 12B models, analyzing the extent to which they can approach the performance of larger 24B and 32B models in the ED selection setting.
\item An analysis of abstention-aware ED in this setting.
\end{enumerate}

Our implementation is publicly available.\footnote{\url{https://github.com/FinaPolat/RAISED}}

\section{Related Work}

Entity Disambiguation (ED) is the final stage of an Entity Linking (EL) pipeline, where ambiguous mentions are mapped to entities in a Knowledge Base (KB). Classical EL pipelines separate candidate generation from contextual disambiguation \cite{ROSALESMENDEZ2020100600}, while recent approaches increasingly rely on pretrained and large language models for retrieval, selection, or both \cite{entityLinkingWikidataSurvey2026,sevgili2022neural}. Here, we focus on recent work with respect to end-to-end and modular entity linking, the use of LLMs for context enrichment and dealing with unretrieved entities. 

\subsubsection{End-to-End and Modular ED.} Recent ED systems commonly adopt either end-to-end neural architectures or modular retrieval-and-selection pipelines. End-to-end approaches such as ReFinED \cite{ayoola-etal-2022-refined}, ExtEnD \cite{barba-etal-2022-extend}, FusionED \cite{wang-etal-2024-entity}, and VERBALIZED \cite{rücker2025evaluatingdesigndecisionsdual} jointly optimize candidate retrieval and disambiguation using pretrained language models, dense retrieval, or encoder-decoder architectures. While these methods achieve strong benchmark performance, they typically depend on trained retrievers and fixed indexes.

In parallel, recent work has explored modular pipelines that use LLMs for contextual selection over externally retrieved candidates. Most modular ED pipelines rely on dense retrievers such as BLINK \cite{wu-etal-2020-scalable} or autoregressive retrieval methods such as GENRE \cite{cao2021autoregressive}. Systems such as InsGenEL \cite{xiao-etal-2023-instructed}, ChatEL \cite{ding-etal-2024-chatel}, EntGPT \cite{ding2025entgptentitylinkinggenerative}, OneNet \cite{liu-etal-2024-onenet}, LELA \cite{haffoudhi2026lela}, and OneGen \cite{zhang-etal-2024-onegen} demonstrate that LLMs can effectively perform candidate selection or reranking, often without task-specific fine-tuning. In our experiments, we situate our framework alongside this line of work and additionally repurpose the dual-encoder component of VERBALIZED as a standalone retriever, allowing us to compare a state-of-the-art trained retriever against sparse and Web-based alternatives within a common modular pipeline. 

\subsubsection{LLMs for Contextual Augmentation.} A complementary line of work uses LLMs to enrich contextual information for ED. LLMAEL \cite{xin2024llmaellargelanguagemodels} generates entity descriptions with an LLM based on candidates retrieved by BLINK or GENRE, while Vollmers et al.~\cite{vollmers-etal-2025-contextual} use LLMs to augment mention context in a joint NER and ED setting. Our work instead focuses on retrieval-selection modularity and uses factual descriptions directly retrieved from Wikipedia or Wikidata.

\subsubsection{Abstention, NIL, and NoC in ED.} Recent work has highlighted that ED systems are often forced to predict an entity even when the correct entity is unavailable. Prior research distinguishes between NIL prediction, where the entity does not exist in the KB \cite{zhu-etal-2023-learn}, and None-of-the-Candidates (NoC) prediction, where the entity exists but is not retrieved \cite{zhou-etal-2024-gendecider}. Van Erp et al.~\cite{van-erp-etal-2016-evaluating} and Rosales-Méndez et al.~\cite{ROSALESMENDEZ2020100600} further discuss how KB incompleteness and benchmark design affect ED evaluation. Our work focuses on the NoC setting in modular ED pipelines, allowing retrieval failures to be analyzed independently from selection behavior.

\section{Methodology}

\begin{figure*}[t]
\centering
\includegraphics[width=\linewidth]{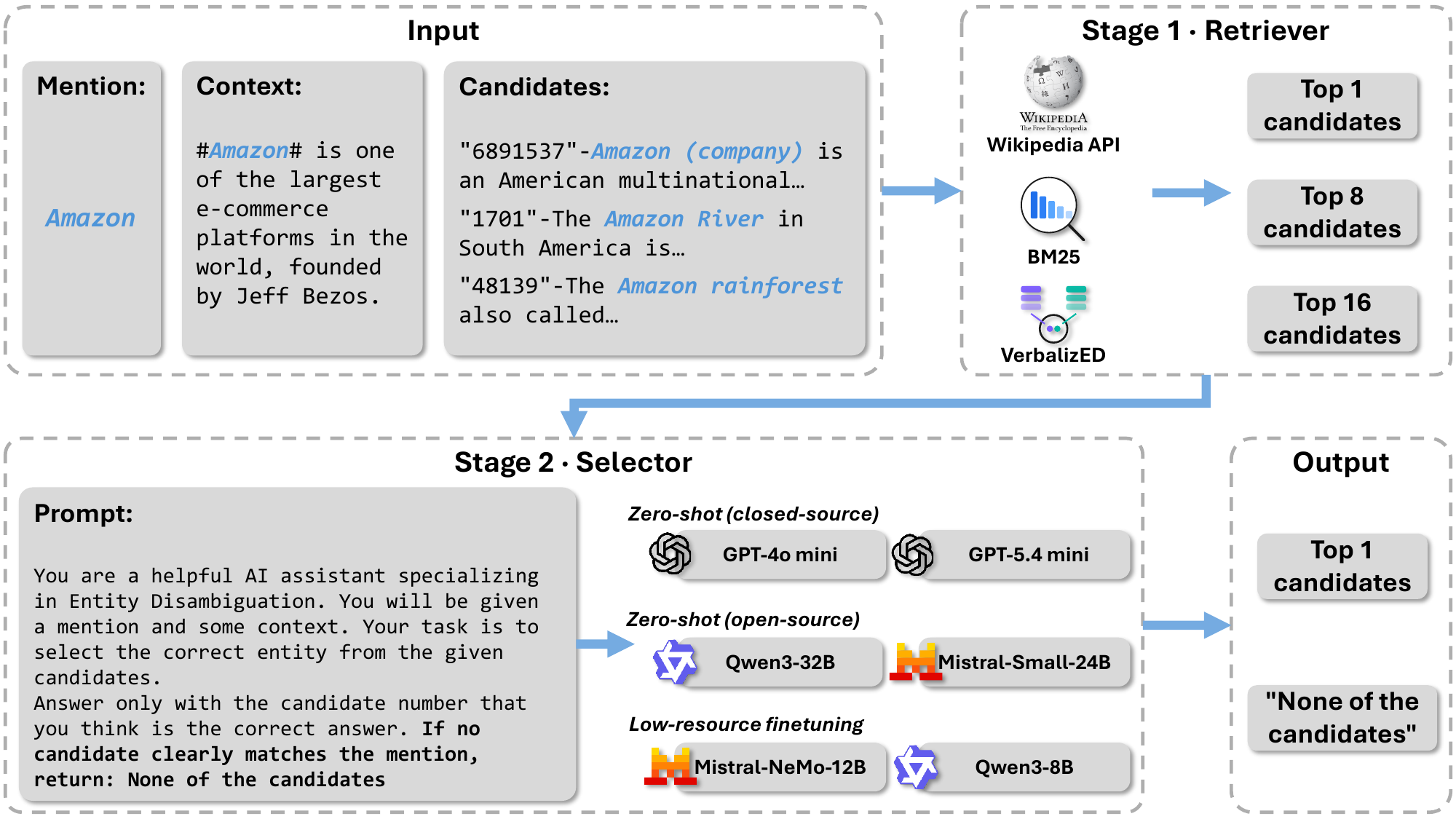}
\caption{Overview of the RAISED framework. Given a mention, its surrounding context, and a large candidate pool from a knowledge base, RAISED operates in two stages. The \textbf{Retriever} ranks entities and returns the top-$16$ candidates; it can be instantiated with sparse, dense, or Web KB search backends. The \textbf{Selector} then prompts an LLM with the mention, surrounding context, and candidate descriptions in a multiple-choice format, and identifies the correct entity or abstains when no candidate is appropriate.}
\label{figure1}
\end{figure*}

\subsection{Problem Statement}

In the ED problem, we are given a textual passage $d = (w_1, \ldots, w_t)$ consisting of a sequence of words, a set of mentions $M = \{m_1, \ldots, m_n\}$ where each $m_i$ is a contiguous span of one or more words in $d$, and a knowledge base (KB, e.g.\ Wikipedia) consisting of a finite set of entities $\mathcal{E}$. The goal of ED is to learn a function
\[
f(d, m_i) \in \mathcal{E} \cup \{\bot\}
\]
that maps each mention to either the correct entity in $\mathcal{E}$ or the abstention symbol $\bot$, indicating that no candidate is appropriate. The abstention case covers two practical situations: NIL, where the correct entity does not exist in the KB \cite{zhu-etal-2023-learn}, and NoC, where the entity exists in the KB but is not surfaced by the retrieval stage \cite{zhou-etal-2024-gendecider}. Later in this work, we focus primarily on the NoC case, as it is the failure mode that arises naturally when retrieval and selection are decoupled.
 \begin{figure}[t]

\small

\fbox{\parbox{0.96\linewidth}{
\textbf{System:} You are an expert entity disambiguation agent. Follow the instructions carefully and do exactly as the user asks. Only provide the output as requested.

\medskip

\textbf{User:}\\

\textit{Task.} Entity disambiguation is the task of identifying which knowledge base entry a marked mention refers to based on its context.

You are given: (1) a text in which the mention is marked as \texttt{\#mention\#}; (2) the mention itself; (3) a list of candidate Wikipedia entities, each with an ID, title, and description.

\textit{Goal.} Select the candidate whose entity best matches the meaning of the mention in the given context.

\textit{Decision guidelines.} Use the surrounding context to determine the intended entity. Prefer the candidate whose description most closely matches the role, domain, or attributes implied by the text. Ignore candidates that share the same name but refer to different concepts. If no candidate clearly matches, output: \texttt{None of the candidates}.

\textit{Output rules.} Output \emph{only} the candidate ID number. No explanations, no punctuation. If no candidate applies, output exactly: \texttt{None of the candidates}.

\smallskip

\textbf{Text:} \texttt{<passage with \#mention\# marked>}\\

\textbf{Entity Mention:} \texttt{<mention>}\\

\textbf{Candidates:} \texttt{<list of (id, title, description) tuples>}\\

\textbf{Answer:}

}}

\caption{Prompt template used by the RAISED selector. The selector is asked to output a candidate ID or the literal string \texttt{None of the candidates} when retrieval has failed. A fully instantiated example is given in Appendix~\ref{app:prompt-example}.}

\label{fig:prompt}

\end{figure}

\subsection{Overview of RAISED}
To experiment with decoupling, we adopt  RAISED (Retrieval And Inference-based Selection for Entity Disambiguation) - a framework for solving the ED problem. We illustrate the architecture in Fig.~\ref{figure1}. RAISED decomposes the function $f$ into two components with distinct objectives: a \textbf{retriever} for candidate generation that produces a high-recall candidate set, and a \textbf{selector} that performs fine-grained, context-aware selection over that set and may also abstain. This decomposition reflects an observation that motivates the paper as a whole: retrieval for candidate generation and selection place different demands on representations --- broad coverage over potentially millions of entities for retrieval, and fine-grained contextual discrimination among semantically similar candidates for selection. By separating the two stages, each component can be specialized to its objective, and the retriever in particular can be lightweight or training-free without compromising end-to-end performance.

\subsection{Retrievers for candidate generation}
\label{sec:retriever}

The retrievers serve as a shortlisting mechanism in RAISED. Therefore, recall is the only objective of the retrieval stage: the retriever's task is to ensure that, whenever the correct entity exists in the KB, it appears in the top-$k$ candidate set $C_i$ passed to the selector. We deliberately keep no architectural commitment at this stage and evaluate three retriever-based candidate generation strategies that span the design space from training-free lexical search to a state-of-the-art trained dense encoder.

All candidate generation strategies considered in this work operate over a Wikipedia-derived entity spaces. Wikipedia API search queries Wikipedia directly, BM25 operates over the Wikipedia-based ZELDA candidate dictionary, and VERBALIZED retrieves from a Wikidata entity set aligned with Wikipedia.

\subsubsection{Wikipedia API search.} The first retriever is the public search interface of Wikipedia, queried directly through its API. Given a mention $m_i$,  the surface form is used as a query and the top-16 matching pages are retrieved. These pages are mapped to KB entries via their Wikipedia identifiers. For each retrieved page, its full abstract is obtained. This is generally too long to include verbatim in the selector prompt. Therefore, we tokenize each abstract into sentences using the NLTK sentence tokenizer~\cite{bird2009nltk} and retain the first three sentences as the candidate description. This retriever is training-free, requires no local index, and benefits from the lexical and fuzzy-matching infrastructure that Wikipedia maintains for its own search functionality. It also reflects the practical scenario in which a system disambiguates against an evolving Web KB without maintaining a local copy.

\subsubsection{BM25 over the ZELDA candidate set.} The second retriever is BM25 \cite{robertson1995okapi} run over the candidate dictionary distributed with the ZELDA benchmark \cite{milich-akbik-2023-zelda}. The dictionary maps surface forms to candidate Wikipedia entities and is shared across all systems trained or evaluated on ZELDA, which makes BM25 over this set a strong, fully training-free baseline directly comparable to the supervised approaches reported by Milich and Akbik \cite{milich-akbik-2023-zelda}. We index entity titles and descriptions and rank candidates by BM25 score against the mention together with a window of surrounding context.

\subsubsection{Repurposed VERBALIZED.} The third retriever is the dual-encoder VERBALIZED~\cite{rücker2025evaluatingdesigndecisionsdual}, which is also used as the primary trained dense retrieval baseline. VERBALIZED is used because prior work demonstrated competitive performance against established ED architectures including GENRE, LUKE, FusionED, FEVRY, and BLINK on the ZELDA benchmark while using substantially less supervision data than BLINK.

VERBALIZED follows the bi-encoder paradigm: a mention encoder produces a contextualized embedding for the mention together with its surrounding text, while an entity encoder embeds KB entries using verbalized Wikidata descriptions. At inference time, the model ranks entities directly by embedding similarity without requiring a separate candidate generation stage.

The original VERBALIZED paper reports that the model compares favorably against the original BLINK bi-encoder despite being trained on considerably fewer instances and entities, although BLINK's full bi-encoder + cross-encoder pipeline remains stronger overall. This makes VERBALIZED one of the strongest available pure dual-encoder ED systems on ZELDA.

For our experiments, we employ the authors' released iterative-training checkpoint with $k=16$ together with their released Wikidata verbalizations and encoders.  We verify that this configuration faithfully reproduces the published VERBALIZED behavior: our recall@1 matches the original scores reported by Rücker and Akbik \cite{rücker2025evaluatingdesigndecisionsdual} to within 0.03 points on average. Casting the same trained model both as an end-to-end ED baseline and as a standalone retriever allows us to isolate the effect of replacing the original dual-encoder decision rule with an LLM-based selector while keeping retrieval fixed.

These three retrievers span the design space we wish to characterize: a Web-based search interface (Wikipedia API), a sparse training-free baseline (BM25), and a trained dense retriever (VERBALIZED). All three feed the same selector, isolating the effect of retriever choice.

\subsection{Selectors}
\label{sec:selector}

The selector takes the passage $d$, the mention $m_i$, and the candidate set $C_i$ produced by the retriever, and returns either an entity $\hat{e}_i \in C_i$ or the abstention symbol $\bot$. We instantiate the selector as an LLM prompted in a multiple-choice format. The prompt enumerates each candidate by ID, title, and KB description, and asks the model to output the ID of the candidate whose entity best matches the mention in context, or the literal string \emph{None of the candidates} if no candidate is appropriate. The full prompt template is shown in Fig.~\ref{fig:prompt}; an example instance is included in Appendix~\ref{app:prompt-example}.

\subsubsection{Zero-shot prompting.} In the zero-shot setting, the selector is an off-the-shelf LLM with no task-specific training. We evaluate four models that span both proprietary and open-source families and a range of capacities: GPT-4o-mini and GPT-5.4-mini from OpenAI, and Qwen3-32B\footnote{\url{https://huggingface.co/Qwen/Qwen3-32B}} \cite{qwen3technicalreport} and Mistral-Small-24B\footnote{\url{https://huggingface.co/mistralai/Mistral-Small-24B-Instruct-2501}} \cite{mistral_small_4_2026} as open-source alternatives. All four models are prompted with an identical template and queried at a low temperature, i.e., 0.01, to reduce sampling variance and facilitate comparison across runs.

\subsubsection{Low-resource fine-tuning.} To examine whether smaller open-source models can also be used as selectors, we additionally fine-tune two models in a low-resource regime: Qwen3-8B\footnote{\url{https://huggingface.co/Qwen/Qwen3-8B}} \cite{qwen3technicalreport} and Mistral-Nemo-12B\footnote{\url{https://huggingface.co/mistralai/Mistral-Nemo-Instruct-2407}} \cite{mistralnemo2024}. 

Fine-tuning uses parameter-efficient adapters with QLoRA \cite{dettmers2024qlora} via the unsloth library\footnote{\url{https://github.com/unslothai/unsloth}}, trained for a single epoch on a single A100 GPU. The training set consists of 1{,}000 examples randomly sampled from the ZELDA training split, with candidates generated by the Wikipedia API; each training instance follows the same multiple-choice prompt format used at inference, with the gold candidate ID as the target. This is intended as an exploratory probe of whether even minimal fine-tuning of smaller 8B--12B models can begin to approach the performance of larger 24B--32B models used zero-shot, rather than a systematic fine-tuning study. All fine-tuned model checkpoints \footnote{\url{https://huggingface.co/FinaPolat/RAISED_QWEN_8B_SFT}} \footnote{\url{https://huggingface.co/FinaPolat/RAISED_Mistral-Nemo_SFT}} are publicly available on Hugging Face.

\subsubsection{Abstention.} Both zero-shot and fine-tuned selectors are given the option to output \emph{None of the candidates}, which we map to the abstention symbol $\bot$ at evaluation time. Section~\ref{sec:evaluation} reports both the standard inKB metrics, in which abstention is treated as a wrong answer, and an abstention-aware metric, in which correct abstentions are rewarded.

\subsubsection{Implementation Details.} Open-source selectors (Qwen3-32B, Mistral-Small-24B, Qwen3-8B, Mistral-Nemo-12B) are run on two NVIDIA A100 GPUs using the vLLM inference library \cite{vllm_paper} with tensor parallelism. Closed-source selectors (GPT-4o-mini, GPT-5.4-mini) are queried through the OpenAI API. All selectors are queried at temperature 0.01 and use the same prompt template (Fig.~\ref{fig:prompt}). Unless otherwise noted, the selector receives $k=16$ candidates from the retriever; recall and selection-accuracy trade-offs at other values of $k$ are discussed in Section~\ref{sec:results-recall}.

\subsection{Datasets}

Evaluation is done using the ZELDA benchmark suite~\cite{milich-akbik-2023-zelda}, which aggregates widely used ED datasets spanning multiple domains and text genres, including newswire, web pages, and social media. ZELDA additionally includes the ShadowLinks benchmark, designed to evaluate robustness under entity overshadowing, i.e., situations in which a less prominent entity is difficult to retrieve or disambiguate because a more popular entity with the same or a similar surface form dominates search results. The suite provides a unified training corpus, standardized entity vocabulary, and predefined candidate sets, enabling controlled comparison across retrieval and selection settings.

\textbf{AIDA-B}~\cite{aida_paper} is a standard newswire ED benchmark derived from Reuters articles in the AIDA-CoNLL corpus. 
\textbf{TWEEKI}~\cite{harandizadeh-singh-2020-tweeki} contains short and informal Twitter mentions with high lexical ambiguity. 
\textbf{REDDIT-POSTS} and \textbf{REDDIT-COMMENTS}~\cite{botzer2021reddit} consist of conversational Reddit discussions annotated for entity linking. 
\textbf{WNED-WIKI} and \textbf{WNED-CWEB}~\cite{wned_cweb_paper} contain entity linking annotations over Wikipedia and general web documents, respectively.

\textbf{ShadowLinks}~\cite{provatorova-etal-2021-robustness} evaluates robustness to entity overshadowing, where highly popular entities compete with contextually correct but less frequent alternatives. It includes three subsets: \textbf{Slinks-Top}, where the correct entity is highly popular; \textbf{Slinks-Shadow}, where the correct entity is overshadowed by a more frequent alternative; and \textbf{Slinks-Tail}, which focuses on low-ambiguity long-tail entities.

Together, these datasets allow us to evaluate retrieval and selection behavior across varying levels of ambiguity, domain shift, and linguistic noise.

\subsection{Evaluation}
\label{sec:evaluation}

Two complementary metrics are used for evaluation: standard inKB micro-F1 and an abstention-aware micro-F1.

\paragraph{InKB micro-F1.}
Following prior ED work, we report inKB micro-F1:
\begin{equation}
\text{micro-F1} = \frac{\text{TP}}{\text{TP} + \frac{1}{2}(\text{FP} + \text{FN})},
\end{equation}
where evaluation is restricted to mentions whose gold entity exists in the KB, following the standard evaluation convention used in EL benchmarking frameworks such as GERBIL \cite{usbeck2015gerbil}. Under this metric, abstentions are counted as errors.

\paragraph{Abstention-aware micro-F1.}
Standard inKB evaluation does not distinguish between selection errors and retrieval failures where the correct entity is absent from the retrieved candidate set. We therefore additionally report an abstention-aware micro-F1 that rewards correct abstentions on None-of-the-Candidates (NoC) cases.

Let $G_i$ denote the gold entity, $\hat{e}_i \in \mathcal{E} \cup \{\bot\}$ the prediction, and $C_i$ the retrieved candidate set. We define: \textbf{TP} (correct entity prediction, $\hat{e}_i = G_i$); \textbf{TA} (true abstention, $\hat{e}_i = \bot$ when $G_i \notin C_i$); \textbf{FP} (wrong entity prediction when $G_i \in C_i$); \textbf{FA} (false abstention, $\hat{e}_i = \bot$ when $G_i \in C_i$); and \textbf{FN} (any wrong entity prediction on NoC cases, $\hat{e}_i \in \mathcal{E} \setminus \{G_i\}$ when $G_i \notin C_i$). Correct predictions and correct abstentions are counted as positive outcomes, while incorrect predictions and false abstentions are penalized:

\begin{equation}
\text{micro-F1}_{\text{abs}} = \frac{\text{TP} + \text{TA}}{\text{TP} + \text{TA} + \frac{1}{2}(\text{FP} + \text{FA} + \text{FN})}.
\label{eq:microf1abs}
\end{equation}

This metric allows retrieval failures to be analyzed separately from selection errors.

\paragraph{Retrieval Recall@$k$.}
For diagnostic purposes, we also report retriever recall@$k$, i.e., the fraction of mentions for which the gold entity appears in the top-$k$ candidates.

\section{Results}
\label{sec:results}

\begin{table*}[t]
\centering
\small
\caption{Comparison of retrieval selection settings (RAISED) and the strongest dual-encoder baseline (VERBALIZED) on the ZELDA benchmark. The VERBALIZED baseline score is taken from \cite{rücker2025evaluatingdesigndecisionsdual}. Values indicate inKB micro-F1 score. RAISED variants are evaluated under three candidate-selection settings (Wikipedia API, BM25, and VERBALIZED). The best score in each column is shown in \textbf{bold}, and the second-best is \underline{underlined}.}
\label{tab:ZELDA_scores}
\resizebox{\textwidth}{!}{%
\begin{tabular}{lcccccccccc}
\toprule
    & \textsc{Aida-} & \textsc{Tweeki} & \textsc{Reddit-} & \textsc{Reddit-} & \textsc{Wned-} & \textsc{Wned-} & \textsc{Slinks-} & \textsc{Slinks-} & \textsc{Slinks-} & AVG \\
Method & B & & \textsc{Posts} & \textsc{Comm.} & \textsc{Cweb} & \textsc{Wiki} & \textsc{Tail} & \textsc{Shadow} & \textsc{Top} & \\
\midrule
\emph{Dual Encoder} \\
VERBALIZED & 88.2 & 78.9 & 92.2 & 88.4 & 71.5 & \textbf{90.8} & 98.2 & 66.3 & 65.9 & 82.3 \\
\midrule
\multicolumn{11}{l}{\emph{RAISED — Wikipedia API candidates}} \\
\multicolumn{11}{l}{\quad Zero-shot inference} \\
GPT-4o-mini       & 85.9 & 85.2 & 93.2 & \underline{93.5} & 73.7 & 72.9 & 98.2 & 55.7 & 77.6 & 81.8 \\
GPT-5.4-mini      & 85.4 & 85.8 & 93.4 & 93.3 & 73.7 & 71.9 & 97.2 & 59.7 & 79.6 & 82.2 \\
Qwen3-32B         & 87.3 & 83.9 & 88.6 & 90.4 & 65.4 & 73.2 & 98.0 & 62.0 & 80.0 & 81.0 \\
Mistral-Small-24B & 85.6 & 84.7 & 91.8 & 92.0 & 72.2 & 51.4 & 98.1 & 55.5 & 78.3 & 78.8 \\
\multicolumn{11}{l}{\quad Low-resource finetuning (QLoRA, 1K examples)} \\
Mistral-Nemo-12B  & 78.9 & 81.4 & 90.9 & 89.8 & 66.6 & 65.3 & 96.9 & 37.3 & 67.6 & 75.0 \\
Qwen3-8B          & 87.5 & 84.1 & 91.7 & 91.0 & 66.9 & 72.5 & 96.7 & 57.4 & 75.9 & 80.4 \\
\midrule
\multicolumn{11}{l}{\emph{RAISED — BM25 candidates}} \\
\multicolumn{11}{l}{\quad Zero-shot inference} \\
GPT-4o-mini       & 88.6 & 84.6 & 93.4 & 93.4 & 76.9 & 75.5 & \textbf{98.3} & 69.7 & 85.0 & 85.0 \\
GPT-5.4-mini      & 89.1 & 85.2 & \textbf{94.3} & \textbf{94.3} & \textbf{78.8} & 75.5 & 97.8 & 74.8 & \textbf{86.8} & 86.3 \\
Qwen3-32B         & 89.4 & 84.3 & 89.6 & 90.8 & 69.0 & 76.8 & 97.9 & 76.1 & \textbf{86.8} & 84.5 \\
Mistral-Small-24B & 87.7 & 84.2 & 92.3 & 92.6 & 75.5 & 57.6 & 97.9 & 70.6 & 84.8 & 82.6 \\
\multicolumn{11}{l}{\quad Low-resource finetuning (QLoRA, 1K examples)} \\
Mistral-Nemo-12B  & 80.6 & 82.0 & \underline{94.0} & 92.7 & 70.7 & 68.7 & 96.9 & 55.8 & 80.3 & 80.2 \\
Qwen3-8B          & 89.5 & 83.7 & 91.1 & 92.5 & 70.0 & 76.0 & 97.2 & 70.3 & 84.3 & 83.8 \\
\midrule
\multicolumn{11}{l}{\emph{RAISED — VERBALIZED candidates}} \\
\multicolumn{11}{l}{\quad Zero-shot inference} \\
GPT-4o-mini       & 90.1 & \underline{86.6} & 92.3 & 92.0 & 77.4 & 85.8 & 97.8 & 80.4 & 84.7 & \underline{87.5} \\
GPT-5.4-mini      & \textbf{92.3} & \textbf{89.0} & 90.8 & 89.9 & \underline{78.7} & \underline{89.0} & 97.8 & \underline{83.8} & 84.8 & \textbf{88.5} \\
Qwen3-32B         & 90.1 & 85.3 & 87.5 & 87.9 & 68.9 & 88.7 & 97.7 & \textbf{84.0} & 84.5 & 86.1 \\
Mistral-Small-24B & 90.1 & 85.1 & 90.0 & 89.3 & 75.6 & 87.2 & 97.4 & 81.5 & 82.8 & 86.6 \\
\multicolumn{11}{l}{\quad Low-resource finetuning (QLoRA, 1K examples)} \\
Mistral-Nemo-12B  & 89.8 & 84.8 & 92.6 & 91.4 & 76.1 & 86.1 & 95.7 & 74.8 & 82.1 & 85.9 \\
Qwen3-8B          & \underline{91.2} & 84.8 & 90.1 & 90.1 & 70.3 & 87.9 & 96.7 & 81.8 & 82.5 & 86.2 \\
\bottomrule
\end{tabular}
}
\end{table*}

Table~\ref{tab:ZELDA_scores} reports inKB micro-F1 scores on the nine datasets of the ZELDA benchmark. We compare various retrieval and selector combinations using RAISED against the dual-encoder baseline VERBALIZED across three retrieval settings (Wikipedia API, BM25, and VERBALIZED as retriever), four zero-shot LLM selectors, and two low-resource fine-tuned selectors.

\subsection{Retrieval--Selection Decoupling}

Across retrieval settings, modular retrieval-selection pipelines generally outperform the end-to-end dual-encoder baseline. The strongest configuration, GPT-5.4-mini with VERBALIZED retrieval, achieves an average micro-F1 of 88.5 compared to 82.3 for VERBALIZED used end-to-end. Competitive performance is also obtained with training-free retrieval: BM25 combined with GPT-5.4-mini reaches 86.3 micro-F1 without requiring a trained retriever.

The improvements are relatively consistent across the evaluated selectors, suggesting that gains are not tied to a single LLM. Overall, the results indicate that separating retrieval from contextual selection can be effective in ED settings where strong LLM selectors are available.

\subsection{Comparison of Retrieval Strategies for Candidate Generation}

The results further suggest that retrieval quality and selection quality contribute differently to overall ED performance. VERBALIZED retrieval yields the strongest average results overall, but BM25 remains comparatively close across datasets and selectors despite requiring no retraining or dense indexing. In contrast, retrieval based on the Wikipedia API performs less consistently, particularly on datasets with higher ambiguity or overshadowing effects.

A likely explanation is that BM25 operates over a more curated and disambiguated candidate space, whereas Wikipedia search frequently returns broader or less informative candidates, including disambiguation pages. This difference appears especially pronounced on WNED-Wiki and ShadowLinks-Shadow. 

Taken together, these findings suggest that lightweight sparse retrieval approaches may remain competitive when paired with strong LLM-based selection.

\subsection{Selector Scale and Low-Resource Fine-Tuning}

Larger zero-shot selectors generally achieve the strongest overall performance. However, smaller models fine-tuned with QLoRA on only 1K examples often approach the performance of substantially larger zero-shot models. For example, Qwen3-8B achieves 83.8 micro-F1 with BM25 retrieval, compared to 86.3 for GPT-5.4-mini under the same retrieval setting.

The results therefore suggest that low-resource fine-tuning can partially compensate for model scale in modular ED pipelines. At the same time, the remaining gap indicates that larger models continue to provide advantages for more challenging disambiguation settings, particularly on overshadowing-focused datasets.

\subsection{Dataset-Level Trends}

Performance is relatively stable across standard news and social-media datasets, including AIDA, Reddit, and Tweeki. Greater variation is observed on the ShadowLinks subsets, which are designed to evaluate robustness to entity overshadowing \cite{provatorova-etal-2021-robustness}. On these datasets, stronger LLM selectors combined with high-recall retrieval generally outperform the dual-encoder baseline, suggesting improved contextual discrimination in ambiguous settings.

In contrast, VERBALIZED retains a slight advantage on WNED-Wiki, possibly reflecting closer alignment between the dataset distribution and the supervision used during VERBALIZED training.

\subsection{Retriever Recall Analysis}
\label{sec:results-recall}

\begin{table*}[htb!]
\centering
\caption{Retriever recall@$k$ on the ZELDA test sets for the Wikipedia API, BM25 over the ZELDA candidate dictionary, and the dual-encoder of VERBALIZED (run with the authors' released code).}
\label{tab:retriever-performance}
\footnotesize
\setlength{\tabcolsep}{3pt}
\begin{tabular*}{\textwidth}{@{\extracolsep{\fill}}l*{9}{c}@{}}
\toprule
& \multicolumn{3}{c}{\bf WIKIPEDIA API} & \multicolumn{3}{c}{\bf BM25} & \multicolumn{3}{c}{\bf VERBALIZED} \\
\cmidrule(lr){2-4} \cmidrule(lr){5-7} \cmidrule(lr){8-10}
\bf DATASET & R@1 & R@8 & R@16 & R@1 & R@8 & R@16 & R@1 & R@8 & R@16 \\
\midrule
\multicolumn{10}{l}{\emph{Datasets in the ZELDA Benchmark}} \\
\bf SLINKS SHADOW & 6.4 & 69.5 & 82.5 & 12.3 & 73.0 & 84.0 & 66.3 & 93.6 & 95.4 \\
\bf SLINKS TOP    & 35.2 & 89.7 & 93.6 & 47.6 & 88.8 & 92.6 & 66.3 & 87.4 & 89.6 \\
\bf SLINKS TAIL   & 96.3 & 99.5 & 99.5 & 94.3 & 98.0 & 98.0 & 98.4 & 99.2 & 99.2 \\
\bf TWEEKI        & 82.0 & 93.6 & 95.0 & 74.1 & 87.2 & 89.0 & 78.3 & 91.7 & 92.6 \\
\bf REDDIT CMMNTS & 95.1 & 98.3 & 98.7 & 91.7 & 96.9 & 97.5 & 88.9 & 95.8 & 97.2 \\
\bf REDDIT POSTS  & 96.4 & 98.6 & 98.8 & 93.6 & 96.3 & 97.2 & 91.2 & 97.6 & 98.2 \\
\bf AIDA          & 74.4 & 93.1 & 94.8 & 65.0 & 89.3 & 92.2 & 87.1 & 97.0 & 97.7 \\
\bf WNED          & 72.0 & 83.9 & 85.8 & 56.3 & 74.2 & 77.4 & 90.4 & 97.7 & 98.3 \\
\bf CWEB          & 71.7 & 87.9 & 90.6 & 56.3 & 80.8 & 85.7 & 71.1 & 85.1 & 87.4 \\
\midrule
\bf AVERAGE       & 69.9 & 90.5 & 93.3 & 65.7 & 87.2 & 90.4 & 82.0 & 93.9 & 95.1 \\
\bottomrule
\end{tabular*}
\end{table*}

Table~\ref{tab:retriever-performance} reports recall@$k$ for the three retrieval strategies on the ZELDA test sets. Recall@$k$ measures the fraction of mentions for which the gold entity appears in the top-$k$ retrieved candidates and provides an upper bound on achievable ED performance for a given retriever-selector pair.

\paragraph{Retriever comparison.}
VERBALIZED achieves the highest recall across all cutoffs, particularly on more challenging datasets such as ShadowLinks-Shadow and WNED. BM25 generally trails the dense retriever, while Wikipedia API retrieval shows more variable behavior across datasets.

At the same time, the recall differences between retrievers decrease at larger candidate sizes. At recall@16, the gap between BM25 and VERBALIZED narrows substantially, which is consistent with the end-to-end ED results: once the gold entity is included in the candidate set, strong LLM selectors can often compensate for moderate retrieval differences.

\paragraph{BM25 and end-to-end performance.}
Although BM25 over a curated candidate dictionary\cite{milich-akbik-2023-zelda} obtains lower recall than VERBALIZED\cite{rücker2025evaluatingdesigndecisionsdual} overall, it remains competitive in end-to-end ED performance. This suggests that candidate quality, rather than recall alone, also affects downstream LLM selection.

\paragraph{Dataset-level trends.}
The largest retrieval differences appear on the ShadowLinks subsets, which evaluate robustness to entity overshadowing \cite{provatorova-etal-2021-robustness}. Here, VERBALIZED substantially outperforms lexical retrieval approaches at low recall cutoffs, indicating the advantage of contextualized dense retrieval in highly ambiguous settings. In contrast, all retrievers perform similarly on ShadowLinks-Tail, where entities are associated with more distinctive surface forms.

Overall, the recall analysis supports the broader finding that retrieval and selection contribute differently to ED performance. Dense retrievers provide advantages in difficult retrieval settings, while lightweight sparse retrieval remains effective when paired with strong LLM-based selection.

\subsection{Abstention-Aware Evaluation}
\label{sec:results-abstention}

\begin{table}[htb]
\centering
\small
\setlength{\tabcolsep}{4pt}
\caption{Abstention analysis: performance when the model outputs ``None of the Candidates'' (NOC). \textsc{NOC-aware F1} is the average inKB micro-F1 across the nine ZELDA datasets when correct NOC predictions are rewarded. \textsc{Gold} is the share of instances where the gold entity is genuinely absent from the candidate set; \textsc{Pred} is how often the model abstains. Precision, Recall, and F1 in the rightmost block are simple means across the nine datasets, computed over gold-NOC instances only.}
\label{tab:abstention}
\begin{tabular}{lc@{\hskip 12pt}cc@{\hskip 12pt}ccc}
\toprule
& NOC-aware & \multicolumn{2}{c}{NOC \%} & \multicolumn{3}{c}{NOC metrics} \\
\cmidrule(lr){3-4} \cmidrule(lr){5-7}
Method & F1 & \textsc{Gold} & \textsc{Pred} & Precision & Recall & F1 \\
\midrule
\multicolumn{7}{l}{\emph{RAISED --- Wikipedia API candidates}} \\
GPT-4o-mini       & 87.5 & 12.1 & 4.9 & 75.9 & 36.5 & 47.3 \\
GPT-5.4-mini      & 87.9 & 12.1 & 3.8 & 70.9 & 29.3 & 38.3 \\
Qwen3-32B         & 85.9 & 12.1 & 7.3 & 60.7 & 38.9 & 43.5 \\
Mistral-Small-24B & 85.0 & 12.1 & 7.1 & 63.5 & 22.0 & 28.7 \\
Mistral-Nemo-12B  & 79.9 & 12.1 & 1.8 & 67.6 & 14.5 & 22.6 \\
Qwen3-8B          & 85.3 & 12.1 & 8.8 & 52.0 & 43.6 & 43.4 \\
\midrule
\multicolumn{7}{l}{\emph{RAISED --- BM25 candidates}} \\
GPT-4o-mini       & 89.5 &  9.6 & 4.8 & 75.0 & 39.9 & 50.3 \\
GPT-5.4-mini      & 90.7 &  9.6 & 5.4 & 75.0 & 45.3 & 53.8 \\
Qwen3-32B         & 88.3 &  9.6 & 8.3 & 63.0 & 51.9 & 53.8 \\
Mistral-Small-24B & 86.1 &  9.6 & 7.0 & 67.9 & 31.0 & 38.9 \\
Mistral-Nemo-12B  & 84.4 &  9.6 & 2.0 & 66.3 & 14.3 & 22.9 \\
Qwen3-8B          & 87.6 &  9.6 & 8.7 & 56.1 & 55.9 & 53.1 \\
\midrule
\multicolumn{7}{l}{\emph{RAISED --- VERBALIZED candidates}} \\
GPT-4o-mini       & 89.1 &  4.9 & 4.4 & 49.4 & 42.9 & 43.0 \\
GPT-5.4-mini      & 90.1 &  4.9 & 3.2 & 59.1 & 35.0 & 41.7 \\
Qwen3-32B         & 87.0 &  4.9 & 6.5 & 40.5 & 43.5 & 39.8 \\
Mistral-Small-24B & 88.6 &  4.9 & 1.1 & 54.4 &  8.2 & 11.8 \\
Mistral-Nemo-12B  & 87.6 &  4.9 & 2.7 & 34.3 & 16.3 & 20.2 \\
Qwen3-8B          & 86.9 &  4.9 & 6.9 & 35.9 & 49.5 & 39.0 \\
\bottomrule
\end{tabular}
\end{table}

Table~\ref{tab:abstention} reports abstention-aware evaluation results across the ZELDA datasets. Note, we do not compare to the VERBALIZED baseline because it is not able to abstain.  In addition to standard ED accuracy, the metric rewards correct ``None of the Candidates'' (NoC) predictions when the gold entity is absent from the retrieved candidate set.

\paragraph{Overall trends.}
All retrieval settings benefit from abstention-aware evaluation, with the largest gains observed for BM25 and Wikipedia API retrieval, where retrieval failures occur more frequently. The strongest configuration, BM25 retrieval with GPT-5.4-mini selection, reaches 90.7 abstention-aware F1, compared to 86.3 under standard inKB evaluation.

The results also reflect the underlying retrieval quality. VERBALIZED retrieval produces the lowest proportion of NoC cases (4.9\%), followed by BM25 (9.6\%) and Wikipedia API retrieval (12.1\%). As a result, abstention provides less additional benefit when stronger retrieval is used. At the same time, the higher recall of VERBALIZED comes with candidate sets that are often more semantically similar, making the downstream selection task comparatively harder despite the lower NoC rate.

\paragraph{Abstention behavior.}
The evaluated selectors differ substantially in their abstention behavior. Larger zero-shot models generally achieve the strongest abstention precision and overall NoC F1, suggesting improved calibration in identifying retrieval failures. At the same time, several models abstain conservatively, leading to relatively low recall on NoC instances.

Interestingly, the fine-tuned Qwen3-8B model matches the abstention F1 of much larger zero-shot selectors when paired with BM25 retrieval (53.1, vs.\ 53.8 for both GPT-5.4-mini and Qwen3-32B), despite trailing them on standard inKB F1.

Overall, the abstention-aware analysis highlights an additional advantage of modular ED pipelines: retrieval failures can be identified explicitly and treated separately from contextual selection errors.

\subsection{Qualitative Error Analysis}

To complement the quantitative results, we examined a sample of 30 disagreement cases from the AIDA test set using a shared LLM selector, which is GPT-5.4-mini, paired with Wikipedia API, BM25, and VERBALIZED. Across these examples, three consistent patterns emerged.

\subsubsection{Candidate quality matters more than recall.} 
Several Wikipedia API and VERBALIZED failures showed that retrieving the gold entity is insufficient when the surrounding candidate set is noisy or semantically overlapping. For example, for the mention \textit{“Asian Cup”}, the VERBALIZED retriever included the gold entity (Q369539, \textit{1996 AFC Asian Cup}) alongside semantically similar candidates like the general \textit{AFC Asian Cup} (Q157894) and various regional variants. Despite the correct entity being present, the selector opted for the broader tournament page (Q157894) due to the dense semantic overlap in the descriptions. Similarly, for \textit{“Japan”}, VERBALIZED provided the correct national team (Q170566) but also included diverse entities like the \textit{Japan national under-23 football team} and various cultural indices. In contrast, BM25 on the curated ZELDA dictionary produced more focused sets, which—even with the same top-$k$ cutoff—reduced the risk of the LLM being distracted by “neighboring” entities.

\subsubsection{Overshadowing is a selection problem, not just a retrieval problem.}
Across the AIDA examples, we observed multiple cases where all three retrievers successfully included the gold entity in $C_i$, but the final prediction still depended on whether the selector could overcome surface-form popularity bias. For instance, in several instances of the mention \textit{“England”}, the candidate sets across BM25 and Wikipedia API included the specific sports-related gold entity (Q31717, \textit{England national football team}), but the selector consistently defaulted to the more globally dominant political entity \textit{England} (Q9316). This occurred even when the context clearly described a sporting match. This confirms that even when retrieval is perfect, the selector’s internal prior (surface-form popularity) can “overshadow” the correct contextual link.

\subsubsection{Abstentions concentrate on plausible-but-wrong candidates.} 
The false abstention patterns highlight how the selector prioritizes precision over guesswork when candidates are poorly differentiated. For the mention \textit{“CHINA”}, both Wikipedia API and BM25 retrievers returned a list of 16 plausible but general entities (e.g., the country, the civilization, or various administrative regions). Because none of the retrieved descriptions (such as for ID 5405) provided a definitive match for the specific, highly localized context of the document, the selector opted for \textit{“None of the candidates”} rather than risking an incorrect match. We saw a identical behavior for \textit{“Asian Cup”} under Wikipedia API retrieval, where the retrieved list (including IDs 250683 and 512257) lacked the granular detail necessary for the selector to confidently distinguish the gold edition from general tournament pages.

\section{Conclusion}
\label{sec:conclusion}

This paper examined candidate retrieval-selection modularity in entity disambiguation with LLM-based selectors. Using the RAISED framework, we compared sparse, dense, and Web-based retrieval strategies under a shared selection stage across the ZELDA benchmark suite.

Training-free modular ED pipelines combined with LLM based selection achieve competitive performance relative to recent fully trained transformer-based approaches. In particular, lightweight sparse retrieval methods such as BM25 remain effective when paired with strong LLM selectors, even in cases where their retrieval recall is lower than that of dense retrievers. Our results also show that abstention-aware evaluation offers additional insight into retrieval failure and selector behavior in modular ED systems.

Beyond the zero-shot setting, we explored low-resource fine-tuning of smaller open-source LLMs using QLoRA. Larger zero-shot models achieved the strongest overall results, and a substantial gap to the fine-tuned 8B--12B models remained on most configurations. The gap was smallest on abstention F1 with BM25 retrieval, where Qwen3-8B matched the larger zero-shot selectors. We view these results as exploratory rather than conclusive evidence that small fine-tuned models can substitute for larger ones.

Overall, the findings suggest that retrieval and contextual selection can be studied and optimized as partially independent components, and that strong ED performance can be achieved without relying exclusively on tightly coupled trained retriever architectures.

\section{Limitations and Future Work}
Our experiments focus on English benchmarks within the ZELDA suite; extending the analysis to multilingual and domain-specific KBs is an important direction for future work. We also considered relatively small fine-tuning sets for the low-resource experiments, and larger-scale supervision may further improve smaller selectors. Finally, while we study abstention in the NoC setting, handling NIL entities that are absent from the KB remains outside the scope of this work.

\section{GenAI Usage}
We used generative AI systems, including ChatGPT and Claude, for language editing, rewriting, and stylistic improvements during manuscript preparation. All research ideas, experimental design, analyses, and scientific conclusions are the work of the authors.

\section{Acknowledgments}

This work is funded by the European Union’s Horizon Europe research and innovation programme within the ENEXA project (grant Agreement no. 101070305). Klim Zaporojets also acknowledges funding from the European Union under the Marie Skłodowska-Curie Actions (Grant Agreement No. 101146515).

\bibliographystyle{splncs04}
\bibliography{references}

\newpage
\appendix

\section{Selector Prompt: Instantiated Example}
\label{app:prompt-example}

Below is a complete example of a prompt issued to the selector. The mention is \emph{CHINA}, taken from a news passage in the ZELDA test set, with candidates retrieved by BM25 over the ZELDA candidate dictionary. Candidate descriptions are truncated for readability; the model receives the full descriptions at inference time.

\begin{quote}\small
\textbf{System:} You are an expert entity disambiguation agent. Follow the instructions carefully and do exactly as the user asks you to do. Do not provide any additional information or explanations. Only provide the output as requested by the user.

\medskip

\textbf{User:}

\textit{Task.} Entity linking is the task of identifying which knowledge base entry a marked mention refers to based on its context.

You are given: (1) a text where the entity mention is marked as \texttt{\#mention\#}; (2) the entity mention itself; (3) a list of candidate Wikipedia entities, each with an ID, title, and description.

\textit{Goal.} Select the candidate whose entity best matches the meaning of the mention in the given context.

\textit{Decision guidelines.} Use the surrounding context to determine the intended entity. Prefer the candidate whose description most closely matches the role, domain, or attributes implied by the text. Ignore candidates that share the same name but refer to different concepts. If no candidate clearly matches the mention, return: \texttt{None of the candidates}.

\textit{Output rules.} Output only the candidate ID number. Do not include explanations, text, or punctuation. If no candidate applies, output exactly: \texttt{None of the candidates}.

\medskip

\textbf{Text:} SOCCER - JAPAN GET LUCKY WIN , \texttt{\#CHINA\#} IN SURPRISE DEFEAT . Nadim Ladki AL-AIN , United Arab Emirates 1996-12-06 Japan began the defence of their Asian Cup title with a lucky 2-1 win against Syria in a Group C championship match on Friday . But China saw their luck desert them in the second match of the group ... \\

\medskip

\textbf{Entity Mention:} CHINA

\medskip

\textbf{Candidates:}
\begin{itemize}\setlength{\itemsep}{2pt}
\item \texttt{5405 -- China:} Officially the People's Republic of China (PRC), is a country...\\
\item \texttt{5760 -- History of China:} The earliest known written records of the history...\\
\item \texttt{197233 -- China Miéville:} British urban fantasy fiction author...\\
\item \texttt{25734 -- Taiwan:} Officially the Republic of China (ROC), is a state in East Asia.\\
\item \texttt{19284336 -- Economy of China:} The economy of China is a socialist market economy...\\.
\item \texttt{1148264 -- China Crisis:} China Crisis are an English pop/rock band.\\
\item \texttt{1863001 -- Bone china:} A type of porcelain composed of bone ash...\\
\end{itemize}
(remaining 9 candidates omitted)\\

\medskip

\textbf{Answer:} 
\end{quote}

The selector receives the full candidate set $k=16$ and outputs only the chosen ID. In abstention cases such as this example, the selector instead outputs the literal string \texttt{None of the candidates}, which is mapped to $\bot$ during evaluation.

\end{document}